%% file: 2020RAL-Lin-bitouch.tex
\documentclass[letterpaper, 10 pt, journal, twoside]{ieeetran} 
\IEEEoverridecommandlockouts                               
\usepackage{etoolbox}
\makeatletter
\patchcmd{\@makecaption}
  {\scshape}
  {}
  {}
  {}
\makeatother

\IEEEoverridecommandlockouts                               

\usepackage{bm}

\usepackage{graphicx}                       
\usepackage{graphics}                       
\usepackage{epsfig}                         
\usepackage[tight,footnotesize]{subfigure}  
\graphicspath{./pics/RAL_fig_for_revise}
\graphicspath{./pics/RAL_new_fig}
\usepackage{amssymb,amsmath}
\usepackage{mdwmath}
\usepackage{commath}   
\usepackage{eqparbox}
\usepackage{mathtools}
\usepackage[utf8]{inputenc} 
\usepackage[english]{babel}
\input{math/rap2texdefs.tex}    

\usepackage{stfloats}                       
\usepackage{url} 
\usepackage{hyperref}
\usepackage{cite}                           
\usepackage[T1]{fontenc} 
\usepackage{tabularx}
\usepackage{diagbox} 
\usepackage{color}
\usepackage{multirow}
\usepackage[table]{xcolor}
\usepackage{longtable}
\usepackage{booktabs} 			
\usepackage{xcolor,colortbl}    

\usepackage{multicol}
\usepackage{graphicx}
\usepackage{placeins}
\usepackage{overpic}
\usepackage{amssymb}
\usepackage{pifont}

\usepackage[fleqn]{nccmath}

\title{Bi-Touch: Bimanual Tactile Manipulation with Sim-to-Real Deep Reinforcement Learning}

\author{
Yijiong Lin, 
Alex Church, 
Max Yang,
Haoran Li,
John Lloyd, 
Dandan Zhang,
Nathan F. Lepora \\
\thanks{
YL was supported by a UoB-CSC-joint scholarship. NL were supported by a Leadership Award from the Leverhulme Trust on ‘A biomimetic forebrain for robot touch’ (RL-2016-39).
{\em (Corresponding author: Yijiong Lin)}
}
\thanks{All authors are with the Department of Engineering Mathematics and Bristol Robotics Laboratory, University of Bristol, Bristol BS8 1UB, U.K. (email: \{yijiong.lin, n.lepora\}@bristol.ac.uk)}
}

\begin{document}
\maketitle
\begin{abstract}

Bimanual manipulation with tactile feedback will be key to human-level robot dexterity. However, this topic is less explored than single-arm settings, partly due to the availability of suitable hardware along with the complexity of designing effective controllers for tasks with relatively large state-action spaces. Here we introduce a dual-arm tactile robotic system (Bi-Touch) based on the Tactile Gym 2.0 setup that integrates two affordable industrial-level robot arms with low-cost high-resolution tactile sensors (TacTips). We present a suite of bimanual manipulation tasks tailored towards tactile feedback: bi-pushing, bi-reorienting and bi-gathering. To learn effective policies, we introduce appropriate reward functions for these tasks and propose a novel goal-update mechanism with deep reinforcement learning. We also apply these policies to real-world settings with a tactile sim-to-real approach. Our analysis highlights and addresses some challenges met during the sim-to-real application, e.g. the learned policy tended to squeeze an object in the bi-reorienting task due to the sim-to-real gap. 
Finally, we demonstrate the generalizability and robustness of this system by experimenting with different unseen objects with applied perturbations in the real world. Code and videos are available at \url{https://sites.google.com/view/bi-touch/}.

\end{abstract}

\begin{figure*}
  \centering
     \includegraphics[width=0.95\linewidth]{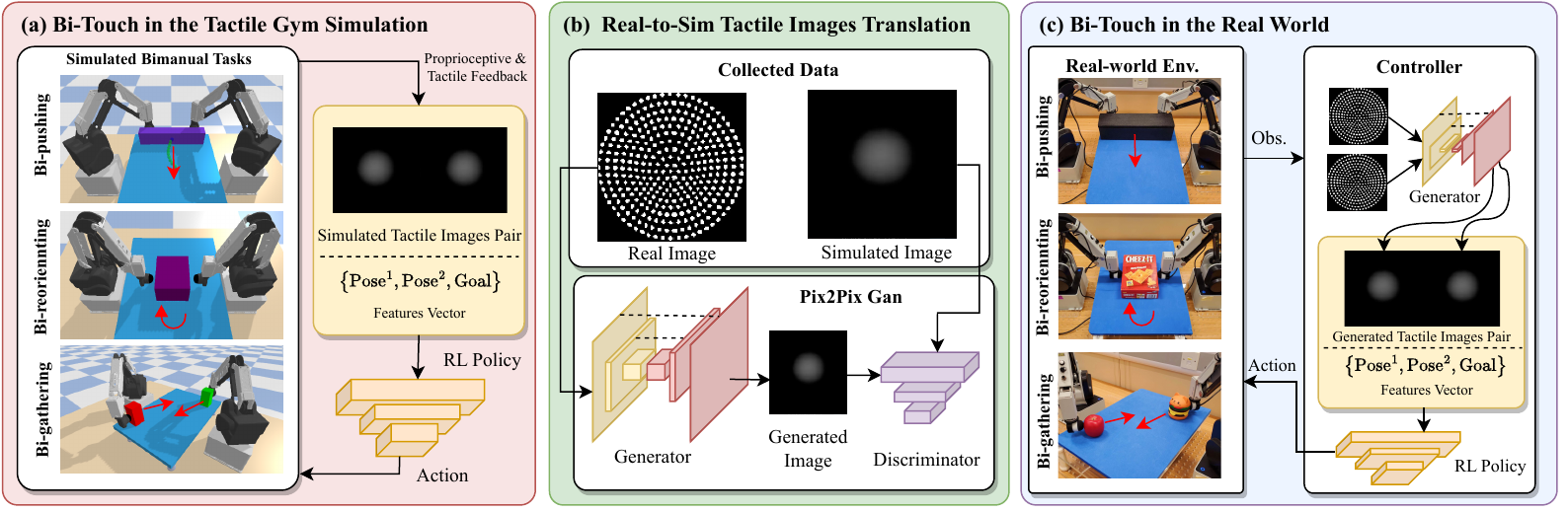}
    \caption{Overview of the proposed dual-arm tactile robotic system (Bi-Touch) with sim-to-real deep RL. a) Deep RL is applied to learn policies for three simulated bimanual tactile manipulation tasks (red arrows show desired displacements) using Tactile Gym. b) Real-to-sim tactile image generator learnt for the surface feature. c)~The real-world evaluation feeds real tactile images through the generator into the RL policy concatenated with proprioceptive information.}
  \label{fig:overview}
  \vspace{0em}
\end{figure*}
\section{INTRODUCTION}\label{sec:Intro}


Bimanual robotic manipulation is a useful and natural way of manipulating large, unwieldy or coupled objects due to the better manoeuvrability, flexibility and a larger workspace compared to single-arm settings \cite{smith2012dual}. Furthermore, the higher dimensionality of the dual-arm state-action space can enable more realistic tasks for real-world application, particularly when tactile sensing is leveraged to complement vision \cite{hansen2022visuotactile}. However, there are challenges in applying bimanual touch: (1)~dual-arm systems often introduce more complexity in terms of system integration and controller design \cite{kruse2014sensor}; and (2)~the high cost of existing dual-arm systems makes them less accessible to the research community. 
Moreover, while vision is commonly used as the primary sensing modality for bimanual manipulation, tactile sensing complements those aspects where vision is limited, such as enabling physically-interactive control with soft contacts and ensuring robustness in scenarios where visual occlusion may occur.



Thus, tactile feedback is needed for precise, safe and reliable dual-arm manipulation, since careful regulation of the applied force on the local contact during robot-object interaction is needed for stable control while avoiding damage~\cite{kruse2014sensor}. However, it remains an open challenge to design effective controllers for dual-arm robots using high-resolution tactile sensing to manipulate unknown objects in uncertain environments~\cite{sintov2020motion}.

Recent advances in deep reinforcement learning (RL) for robotics indicate a plausible route to acquire sophisticated control policies for manipulation tasks with high-dimensional state-action space \cite{james2019sim, andrychowicz2020learning, church_tactile_2021}. Nevertheless, progress in data-driven methods for bimanual robotics is challenging because of the sim-to-real gap, research reproducibility and inaccessible robotic hardware. Also, learning tactile dual-arm manipulation in the real world is difficult, such as preventing the arms from colliding, which is especially difficult at the start of training. Moreover, RL requires frequent manual resetting of tasks which is too laborious in the real world. Lastly, large-scale training in the real world risks damage to sensors. Therefore, the present study aims to advance recent progress with sim-to-real tactile deep RL methods applied to low-cost high-resolution tactile sensors with an affordable dual-arm manipulation system. 


The main contributions of this work are as follows:\\
\noindent1) We adapt and extend the Tactile Gym 2.0 \cite{lin2022tactilegym2} to a low-cost dual-arm tactile robot setting with three new contact-rich bimanual tasks: bi-pushing, bi-reorienting, and bi-gathering.\\ 
\noindent2) We introduce appropriate reward functions for these tasks and show that deep RL reaches satisfactory performance using only proprioceptive and tactile feedback. To improve the robustness of the policies for real-world applications, we improve the sim-to-real transfer for bi-reorienting and propose a novel goal-update mechanism (GUM) for bi-gathering.\\
\noindent3) We demonstrate that the bimanual policies learned in simulation can be transferred well to the physical dual-arm robot. We further demonstrate the generalizability and robustness of the learned policies by testing the system on unseen objects. 

To the best of our knowledge, this is the first framework for sim-to-real deep RL tailored to bimanual tactile manipulation. 

\section{RELATED WORK} \label{sec:related}

\subsection{Bimanual Robot System with Deep RL} \label{subsec:bi_drl_related}

Designing an effective controller for dual-arm manipulation is a long-standing challenge \cite{zhang2019leveraging, garcia2020benchmarking}, and the recent success of deep RL incentivizes robotics researchers to apply it to this problem \cite{ chiu2021bimanual, amadio2019exploiting}. Previous work \cite{kataoka2022bi} successfully applied an RL policy learned in simulation to a real-world task (connecting magnets) with a dual-arm platform, but it relied on a marker-based visual tracking system and did not show any generalization ability to unseen objects. Grannen et al. \cite{grannen2022learning} proposed an RL framework to learn bimanual scooping policies for food acquisition and demonstrated its generalizability on various unseen food.

Since deep RL relies on large amounts of training, in recent years several bimanual manipulation simulators have been developed. Fan et al. developed a distributed RL framework called SURREAL for a set of dual-arm manipulation tasks \cite{fan2018surreal}. Similarly, Zhu et al. \cite{robosuite2020} designed a suite of single-arm and dual-arm manipulation environments with Mujoco for algorithm development and evaluation. Chen et al. \cite{chen2022towards} developed a simulation benchmark for bimanual dexterous hands with a suite of bimanual manipulation tasks and tried solving these tasks with different RL methods. However, these benchmarks have not considered tactile sensing, which hinders their application to tasks that require direct, detailed information about the local contact for fine-grained manipulation. 

\subsection{Bimanual Robot System with Tactile Sensing} \label{subsec:bi_tactile_related}
Tactile sensing enables dual-arm robots to perceive the local contact features with a granularity that cannot be achieved exclusively with vision. Nevertheless, this topic is not widely studied. Sommer et al. \cite{sommer2014bimanual} leveraged a dual-arm tactile robot for object exploration with a Gaussian-Process-based filter and grasp-pose selection using a Gaussian mixture model; the method required human demonstrations for particular objects and only one arm had tactile sensing. Hogan et al. \cite{hogan2020tactile} developed dual-arm controllers with tactile palms for pusher-slider manipulation that can explicitly control the object's trajectory. A drawback of this approach was that it relied on a set of pre-designed motion skills and assumed full knowledge of the environment, hindering its ability to generalize to uncertain environments where unseen objects and unpredictable perturbations may be encountered. Here we aim to facilitate research on bimanual tactile robotics by developing a dual-arm tactile robot platform as a benchmark to design sophisticated controllers for complex bimanual tasks. 
\section{Methods} \label{sec:method}

\subsection{Accessible Dual-arm Tactile Robotic System}\label{subsec:hardware}

\subsubsection{Desktop Dual-arm Platform} To facilitate affordable automation and lower the entry barrier, we develop a low-cost dual-arm tactile robotic system while keeping high accuracy, which is comprised of two industry-capable desktop robotic arms (Dobot~MG400) with vision-based tactile sensors mounted at the wrists as end-effectors (Fig. \ref{fig:overview}c). The proposed platform is developed with the Tactile Gym 2.0 \cite{lin2022tactilegym2} simulation (Fig.~\ref{fig:overview}a) for deep RL-based policy training (see Sec. \ref{subsec:method_task}). 

Although the Dobot~MG400 has a 4-DoF workspace with only Cartesian positions and rotations around the $z$-axis of the end-effector actuated, its accuracy is the same as larger industrial robot arms such as the UR5 at a considerably improved cost and convenience \cite{lin2022tactilegym2}. To maximally leverage the workspace of two robots while having one fixed configuration suitable for all the bimanual tactile robotic tasks considered here, we introduce a table (120\,mm height) into the physical dual-arm platform to support objects, placed between the two arms (shown in blue in Fig.~\ref{fig:overview}a,c). The two arm bases are set centrally below the board separated by a distance of 700\,mm.


\subsubsection{High-resolution optical tactile sensing} To endow the dual-arm robot with tactile sensing, we equip it with two low-cost, high-resolution biomimetic optical tactile sensors used in previous tactile robot research with the Dobot MG400 (the TacTip)~\cite{lepora2022digitac} as the end-effectors. The sensor features an internal array of biomimetic markers on protruding pins inside the soft tactile skin, and the sensing principle of this tactile sensor is to capture marker-based movement that amplifies the skin deformation induced by physical contact against external stimuli. We refer to references~\cite{ward2018tactip,lepora2021soft} for more details. 

\subsection{Sim-to-Real Deep RL Framework for Bimanual Tactile Robotic Manipulation}\label{subsec:TS2R}

To apply the deep RL policies learned in simulation to the physical dual-arm tactile robotic system, we take a sim-to-real approach \cite{church_tactile_2021} consisting of three parts (shown in Fig.~\ref{fig:overview}): \textbf{1)} An online agent training in simulation (Fig.~\ref{fig:overview}a), where deep RL policies are learned in the Tactile Gym for three bimanual tactile robotic tasks (bi-pushing, bi-reorienting and bi-gathering) with observations comprising simulated tactile images and proprioceptive feedback. \textbf{2)} A real-to-sim domain adaption process where a translation model is learned to transfer real to simulated tactile images. \textbf{3)} A sim-to-real application with networks trained in the previous two parts, for transferring deep RL policies to the physical system.

To apply this approach to the Bi-Touch, we make several changes. First, we develop a simulated dual-arm tactile robot in the configuration described above (Sec. \ref{subsec:hardware}) in Tactile Gym that is suited to the three bimanual robotic tasks. In the simulation learning phase, we concatenate two simulated tactile images (depicted in Fig.~\ref{fig:overview}a) to be used as part of the observation of an RL agent. The simulated tactile images are captured by synthetic cameras embedded within simulated tactile sensors built from CAD models of the real sensors.

In the real-to-sim domain adaption phase, a real tactile image dataset paired with a simulated tactile image dataset is required for training. All three bimanual tasks considered here need to model a specific contact feature: the shape of a flat (or relatively flat) contact surface. We adopt the procedure proposed in \cite{church_tactile_2021} for dataset collection. However, a larger sensing space is necessary, as the bimanual tasks require the robot to learn more difficult control than for the previous single-arm tasks~\cite{church_tactile_2021,lin2022tactilegym2}.
Thus we collect surface-feature data with depth range $[0.5,8]$\,mm and rotation range $[-30^{\circ},30^{\circ}]$, broader than in ~\cite{church_tactile_2021,lin2022tactilegym2}. The training dataset comprises 5000 tactile images and the validation dataset has 2000 tactile images collected both in simulation and with a desktop tactile robot on paired random contacts. Samples are labelled with the relative poses between the sensor and a known flat surface. Finally, we use an image-to-image translation Generative Adversarial Network (GAN) \cite{isola2017image} to learn the real-to-sim tactile image translation, with hyperparameters from \cite{church_tactile_2021}.

Since the RL policies take concatenated tactile images as part of the observation, in the sim-to-real application phase (Fig.~\ref{fig:overview}c), we first transfer the real tactile images from both TacTips separately into simulated ones. These tactile images were then concatenated as observation input for the RL agent.

\subsection{Bimanual Tactile Manipulation Tasks}\label{subsec:method_task}

In this study, we propose three bimanual tactile control tasks to benchmark the aforementioned dual-arm tactile system: bi-pushing, bi-gathering and bi-reorienting. The action space of the dual-arm robot in the bi-reorienting task is 6-dimensional comprising the $x$-, $y$-position and $Rz$-rotation angle of each robot's tool-centre-point (TCP), while the ones in the bi-pushing task ($x$ and $Rz$) and the bi-gathering task ($y$ and $Rz$) are 4-dimensional. Note that each end-effector (TacTip) is controlled and moves in its own TCP frame.
\subsubsection{Bi-Pushing} 
\begin{figure}[t]
  \centering
    \includegraphics[width=0.95\linewidth]{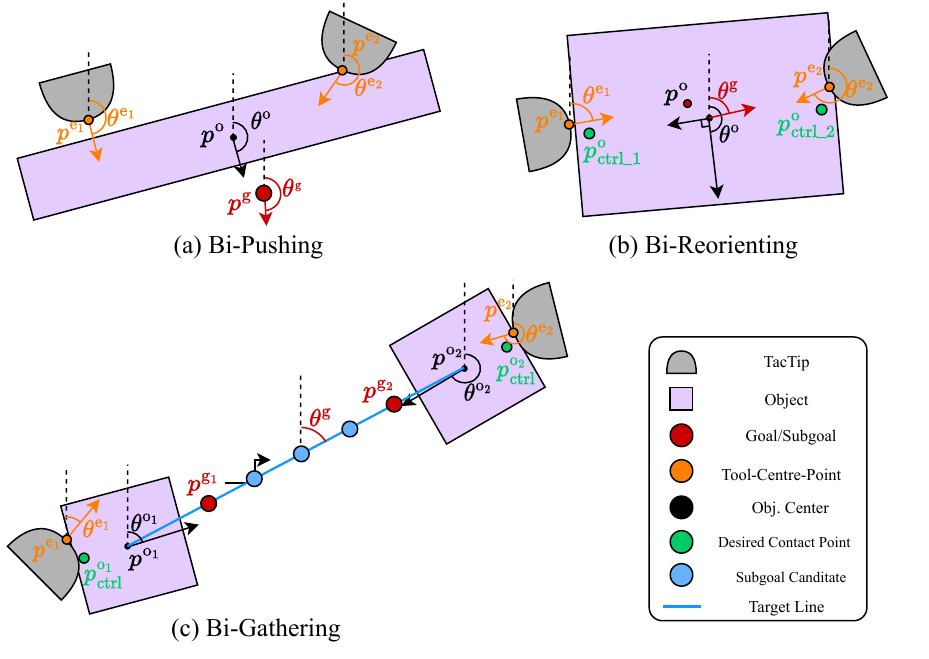}
    \caption{Illustration of the reward functions for all three proposed bimanual tasks: (a) bi-pushing, (b) bi-reorienting, 
 and (c) bi-gathering. The notations in red are the goals (or subgoals) and those in orange are the proprioceptive information that is part of the observation. 
 }
  \label{fig:tasks_ill}
  \vspace{0em}
\end{figure}

An advantage of dual-arm robots over single-arm robots is that they can move relatively large and unwieldy objects. 
The goal of this bi-pushing task is to move a large-size object on a planar surface collaboratively with two robot arms with end-effectors (TacTips) to achieve a sequence of goals along a given trajectory. 
Each goal comprises the target object's position $p^{\text{g}}(x,y)$ and orientation $\theta^{\text{g}}$. We express the reward $R_{t}^{\rm \text{BP}}$ at time step $t$ for the bi-pushing task as:
\begin{equation} 
R_{t}^{\rm \text{BP}} = -w_1\left \| p_{t}^{\rm g} - p_{t}^{\rm o} \right \|_2 -w_2S(\theta^{\rm g}_{t},\theta_{t}^{\rm o})-w_3\sum_{i=1}^{2}S(\theta^{{\rm e}_{i}}_{t},\theta_{t}^{\rm o}),
\label{eq:push}
\end{equation}
where $w_{j}>0$ $(j\in\{1,2,3\})$ are reward weights; $p_{t}^{\rm g}$ and $\theta^{\rm g}_{t}$ are the position and orientation of the current goal respectively; $p_{t}^{\rm o}$ and $\theta^{\rm o}_{t}$ are the current position and orientation of the object; $\theta^{{\rm e}_{i}}_{t}$ is the current orientation of the TCP ${\rm e}_{i}$ for robot arm $i\in\left\{1,2\right\}$; $S(\phi,\psi)=1-\cos(\phi-\psi)$ is the cosine distance between the angles $\phi$ and $\psi$. All notation is illustrated in Fig.~\ref{fig:tasks_ill}a. We interpret the first and second terms of Eq.~(\ref{eq:push}) as encouraging the robot to push the object towards the goal, while the third term is to encourage the robot to maintain the TacTips normal to the contact surface for stable pushing.

\subsubsection{Bi-Reorienting} Reorienting an object with two arms is necessary when the object size exceeds the limit of what can be held by a gripper or a robot hand. The goal of this bi-reorienting task is for two robotic arms to reorient an object located at the workspace centre to a given target angle $\theta^{\rm g}$ while keeping the object centre fixed in place. The dual-arm robot should reorient the object with gentle contact while keeping the end-effectors (TacTips) normal to the contact surface. 
We express the reward $R_{t}^{\rm \text{BR}}$ at step $t$ for the bi-reorienting task as
\begin{align}
R_{t}^{\rm \text{BR}} &= -w_1\left \| p^{\rm o}_{0} - p_{t}^{\rm o} \right \|_2 \label{eq:2_a} -w_2S(\theta^{\rm g},\theta_{t}^{\rm o}) \nonumber\\ 
- & w_3\sum_{i=1}^{2} S(\theta^{{\rm e}_{i}}_{t},(-1)^{i}(\pi/2 + \theta^{\rm o}_{t}))  -w_4 \sum_{i=1}^{2} \left \| p_{\text{ctrl}\_i}^{\rm o} - p_{t}^{e_i}  \right \|_2
\end{align}
where $w_{j}>0$ $(j\in\{1,2,3,4\})$ are reward weights; $p^{\rm o}_{0}$, $p^{\rm o}_{t}$ and $\theta^{\rm o}_{t}$ are the initial/current positions and orientation of the object respectively; $p_{t}^{{\rm e}_{i}}$ and $\theta^{{\rm e}_i}_{t}$ are the current position and orientation of each TCP ${\rm e}_{i}$ ($i\in\{1,2$\}) respectively; $ p_{\text{ctrl}\_i}^{\rm o}$ ($i\in\{1,2$\}) are the desired contact points to either side of an object that are used to specify the desired positions of the TCPs when in contact with the object. All notation is illustrated in Fig.~\ref{fig:tasks_ill}b. The second term in Eq.~(2) encourages the robot to reorient the object to the target angle $\theta^{\rm g}$ while maintaining the object at its original position with the first term. The third term encourages the robot to maintain the TCP normal to the contact surface for stable reorienting. The final term encourages the dual-arm robot to keep both TCPs close to the desired contact points, which helps maintain the contact, especially the contact depth, of the TCP to avoid losing contact with the object. 

However, a challenge we met during the sim-to-real application for this task is that a policy trained using Eq. (2) with the original simulated TacTip dynamics (stiffness, damping, etc) tended to squeeze the object when the goal angle is large in the real world, and this phenomenon became worse as the object length increased (see accompanying video). We attribute this phenomenon to a sim-to-real gap where the dual-arm robot tried to finish the task quickly for maximum rewards, but then learned to rotate the object by squeezing it in simulation to achieve this. However, in the real world, the TacTips may break if over-deformed and the objects are not always of high stiffness, so this learned policy is not suitable. 

To solve this problem, we made several changes for policy learning in the simulation: 1) we tuned the simulated TacTip skin stiffness and damping coefficients to make it more elastic; 2) the dual-arm robot needed to maintain the object in the same pose for 10 time steps to achieve the goal orientation; 3) we increased the penalty coefficient if the dual-arm robot over-squeezes the object (large contact depth); 4) we trained the policy with a higher probability of larger goal angles. After conducting an ablation study for the above changes, we found that change 1) contributes most to the successful sim-to-real application, as the higher stiffness and lower damping make the TacTip skin more elastic against external contact, which enables the policy to be more sensitive to the contact depth as the tactile image are more responsive to depth in simulation.
\subsubsection{Bi-Gathering} Gathering objects together is a common behaviour in our daily life, from tidying our desks to moving and sorting packages in warehouses. The goal of this bi-gathering task is for the dual-arm robot to gather two objects together by pushing them towards each other on a planar surface. Thus, each end-effector of the dual-arm robot has to push an object towards a dynamically changing goal (as the goal position of each object is the other object's current position), 
which makes it a harder exploration problem for RL as compared to single-arm pushing~\cite{lin2022tactilegym2} or bi-pushing (above) that have static goals. The task is considered completed when the distance $d$ between the two objects becomes closer $d<\epsilon$ than a distance threshold $\epsilon$ (related to the size of the objects). We express the reward $R_{t}^{\rm \text{BG}}$ at $t$ of the bi-gathering task as
\begin{align}
R_{t}^{\rm \text{BG}} = -w_1\left \| p^{\rm o_1}_{t} - p_{t}^{\rm o_2} \right \|_2 - w_2\sum_{i=1}^{2}S(\theta^{{\rm e}_{i}}_{t},\theta^{{\rm o}_i}_{t}) \hspace{3em}\nonumber\\
 -w_3\sum_{i=1}^{2}\left \| p_{\rm ctrl}^{{\rm o}_i} - p_{t}^{{\rm e}_{i}} \right \|_2,
\end{align}
\noindent where $w_{j}>0$ $(j=\{1,2,3\})$ are reward weights; $p_{t}^{{\rm o}_i}$  and $\theta^{{\rm o}_i}_{t}$ are the current position and orientation of the object ${\rm o}_i$; $p_{t}^{{\rm e}_{i}}$ and $\theta^{{{\rm e}_{i}}}_{t}$ are the current position and orientation of the TCP ${\rm e}_{i}$; $p_{\rm \text{ctrl}}^{{\rm o}_i}$ specifies the desired contact point position on the object ${\rm o}_i$ for controlling the contact depth of the TCP ${\rm e}_{i}$ ($i\in\{1,2\}$). All notation is illustrated in Fig.~\ref{fig:tasks_ill}c. The first term of Eq.~(3) pushes two objects closer while the second term tries to maintain the TCP normal to the contact surface of the object for stable pushing. The third term encourages the robot to keep the TCPs close to the contact points as much as possible to avoid losing contact.

To further explore the limit of the dual-arm tactile robot, we also introduce random perturbations to the objects during the gathering. Specifically, a random force is applied to an object's centre of mass at a random time step when training. 

\begin{figure*}[t]
  \centering
     \includegraphics[width=0.95\linewidth]{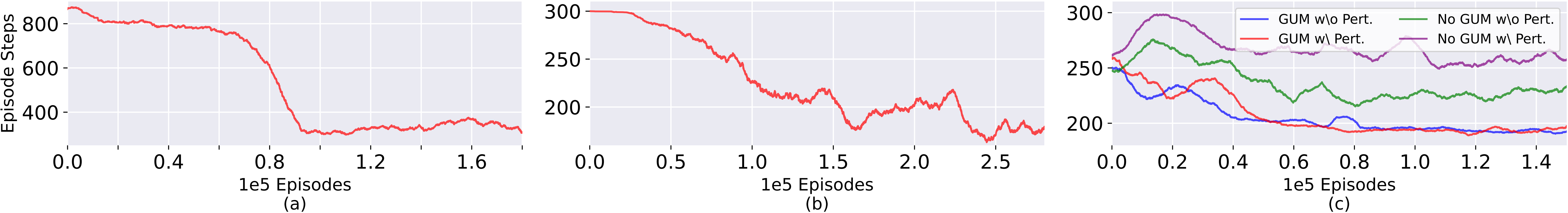}
\vspace{-1em}

    \caption{Task completion times during training for learned policies averaged over 10 trials (a) bi-pushing, (b) bi-reorienting, and (c) bi-gathering tasks. Note that less task completion time means better performance. Our proposed goal-update mechanism (GUM) in the bi-gathering task resulted in improved performance: the blue and red plots (with GUM) show reduced numbers of time steps required to achieve the task in one episode, under both perturbed and unperturbed conditions, compared to the green and purple plots (without GUM).}
  \label{fig:rl_results}
  \vspace{-.5em}
\end{figure*}

Although the most obvious formulation would be to place the goal of an object at another object, we found that this did not work well for a moving sparse goal both without and with perturbations, resulting in suboptimal performance as shown in Fig. \ref{fig:rl_results}c (green and purple plots). Thus, we propose a novel goal-update mechanism (GUM) that can generate subgoals to help the robot learn the task with auxiliary reward signals. Specifically, we generate $n$ static points at equal interval distances along a target line as subgoals for the two objects (as shown by the blue line in Fig.~\ref{fig:tasks_ill}c). We have experimented it with $n=[5, 10, 20]$ and found $n=10$ performed slightly better than the others. The target line is updated every $h$ time steps, and the subgoal location for each robot arm is set to the generated point nearest to the object that is closer to it. We express the reward $R_{t}^{\rm \text{BG-GUM}}$ with GUM as
\begin{align}
R_{t}^{\rm \text{BG-GUM}} = R_{t}^{\rm \text{BG}} - w_4\sum_{i=1}^{N}\left \| p^{{\rm g}_{i}}_{t} - p_{t}^{{\rm o}_i} \right \|_2  \hspace{6em}\nonumber\\
 -w_5\sum_{i=1}^{N}S(\theta^{{\rm o}_i}_{t},(-1)^{i}\theta^{\rm c}_{t}),
\end{align}
\noindent where $w_{j}>0$ $(j=\{4,5\})$ are reward weights; $p^{{\rm g}_i}_{t}$ and $\theta^{\rm c}_{t}$ represent the position of subgoal ${\rm g}_{i}$ for object ${\rm o}_i$ at time step $t$ respectively. All notation is illustrated in Fig.~\ref{fig:tasks_ill}c. The second term of Eq.~(4) guides objects towards the nearest subgoals on the target line, providing denser auxiliary rewards for $R_{t}^{\rm \text{BG}}$. The final term encourages object to be pushed along the target line direction. The hyperparameter $h$ controls the target line update rate, which should update with appropriate frequency to provide sufficient guidance for the robot to move objects in the desired direction, but also avoid updating too often, which can cause the robot to fail to reach the subgoals due to insufficient time steps. We have experimented it with $h=[25, 50, 75, 100]$ and found $h=75$ performed the best.


For the sim-to-real application in this task, since the object pose is unknown in real-world experiments, we instead select the current tool-centre-points (TCPs) for constructing the target line. However, during training in simulation, we found that training from scratch with a line between TCP positions failed to learn. This is because the initial random policy cannot maintain contact with objects and accordingly, this line cannot provide useful information for learning. To circumvent this, we devised a 2-step curriculum in the simulation where we first use objects' centres to construct the target line to train a policy from scratch, and then switch to TCPs for extra training so that the policy can be attuned to the real-world setting. We found that this simple curriculum was critical for successful training in simulation and sim-to-real application. During the curriculum, we also increased the probability of perturbation and its magnitude to further improve the policy's robustness when applied in the real world. 
%
\section{Experiments and Results}\label{sec:exps_results}
\begin{figure}[t]
  \centering
    \includegraphics[width=1\linewidth]{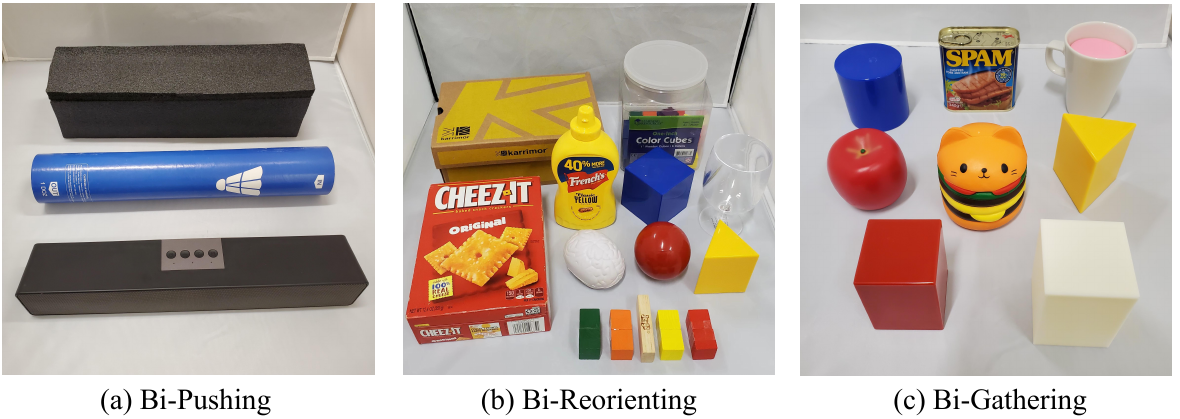}
    \caption{The objects used in the three proposed dual-arm manipulation tasks for real-world testing: a) a tripod box, a shuttles tube, and a loudspeaker for the bi-pushing task; b) a plastic cube, a cracker box (red), a shoe box (yellow), a cubes box (transparent), a mustard bottle, a goblet, a plastic ball, a soft brain toy, a plastic triangle prism, and a set of bricks for the bi-gathering task; c)~a blue cylinder, a spam can, a ceramic mug, an apple, a foam toy, a triangle prism, and two cubes for the bi-gathering task. These objects are unseen during training and are chosen to vary in size, weight, shape and stiffness.}
  \label{fig:real_objs}
  \vspace{0em}
\end{figure}
\begin{figure}[b]
\vspace{-1em}
  \centering
  \includegraphics[width=1\linewidth]{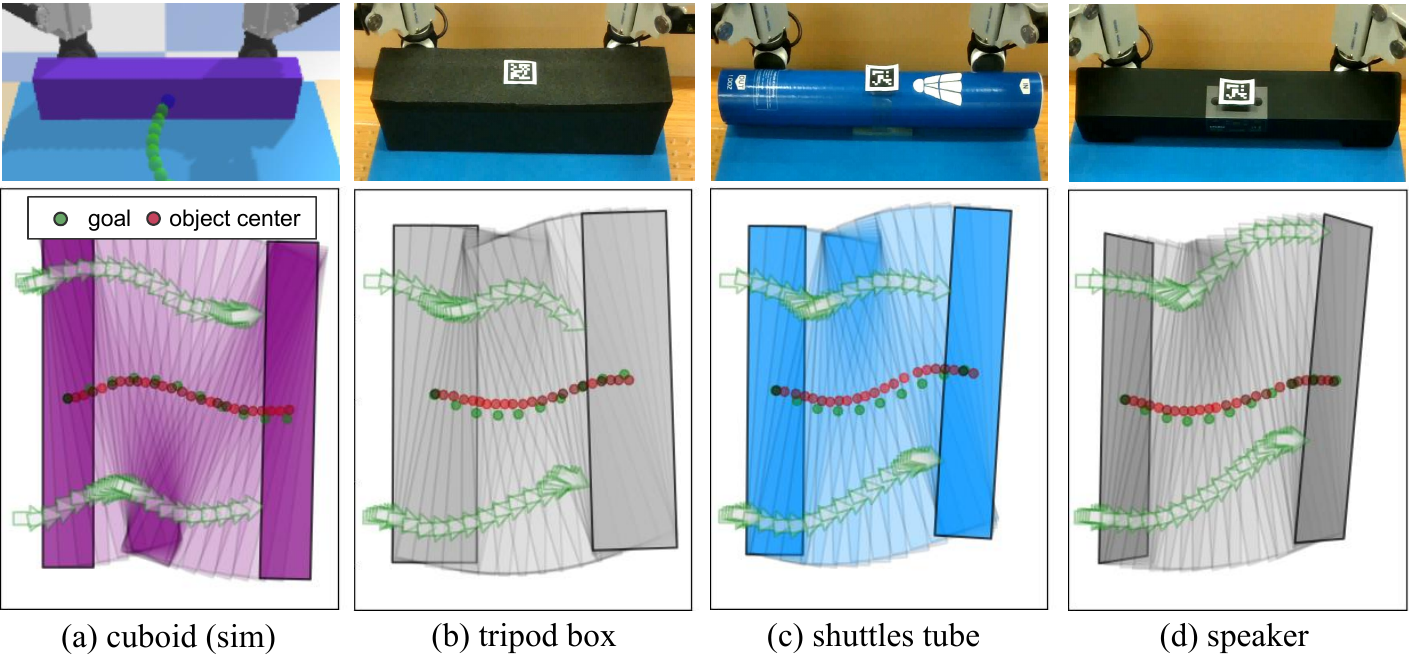}
    \caption{The objects' trajectories plots (in red) of the bi-pushing task with (a) a cuboid (simulation), (b) a tripod box, (c) a shuttles tube and (d) a loudspeaker. The TacTips' trajectories are indicated with green arrows.} 
  \label{fig:push_exp_plots}
\end{figure}
\begin{figure*}[ht]
  \centering
  \includegraphics[width=0.95\linewidth]{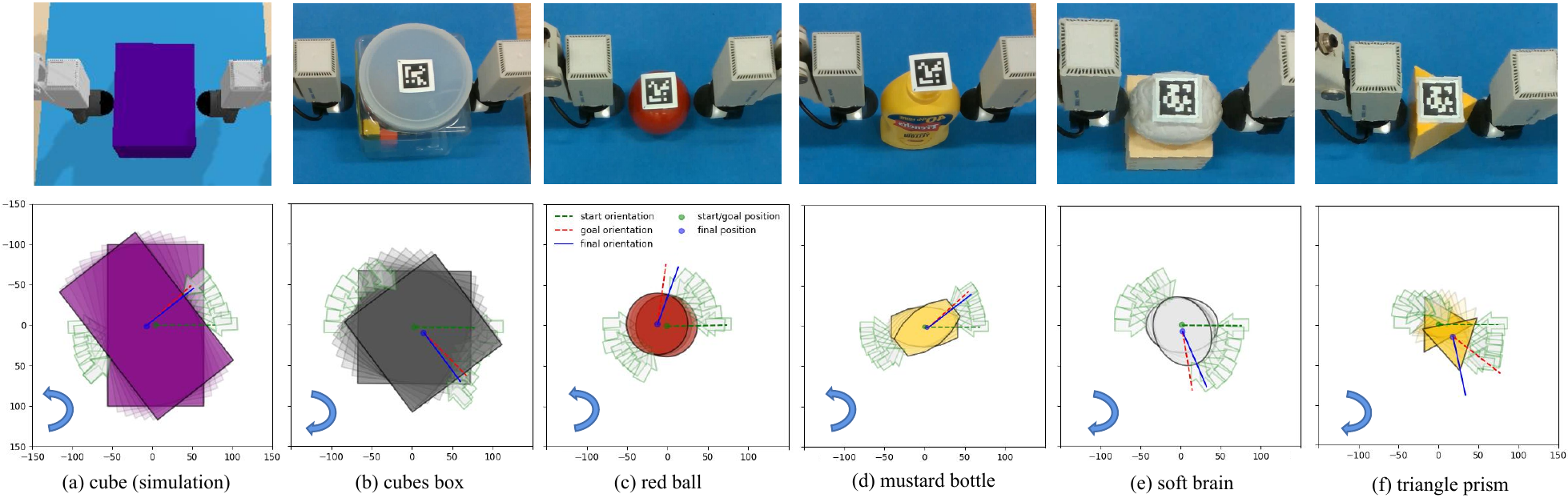}
  \vspace{-1em}
    \caption{Object trajectories for the bi-reorienting task for the (a) cuboid (simulation), (b) cubes box, (c) plastic ball, (d) mustard bottle, (e) soft brain toy and (j) triangular prism. End-effector positions during the trajectories (green arrows) are shown, and the start, goal and final object orientations (lines).} 
  \label{fig:reorient_exp_plots}
  \vspace{-1em}
\end{figure*}
\subsection{Evaluations on the Bi-Touch in Simulation}
An on-policy model-free deep-RL algorithm called Proximal Policy Optimization (PPO)~\cite{schulman2017proximal} is used to train policies in simulation for all three bimanual tactile robotic tasks described above. Specifically, we use the Stable-Baselines-3 \cite{raffin2019stable} implementation of PPO for learning the policies. 

We obtained successfully trained policies in simulation for all three tasks (training curves in Figs~\ref{fig:rl_results}a-c respectively). The bi-pushing is the easiest task to learn with a smooth learning curve and convergence at an early time step. The other two tasks have subtleties in learning that we describe below.

\subsubsection{Bi-Pushing}
In each episode, we selected a sequence of goals from a sampled linear or sinusoidal path as a desired object trajectory (parameterised by $y = kx$ and $y = a\sin(x/50)$ respectively, where $k\in[-0.28,0.28]$, $a\in[-20,20]$ and $x\in[-280,50]$\,mm). The simulated dual-arm tactile robot can learn to push a large object through a trajectory with a training curve shown in Fig. \ref{fig:rl_results}a. A successful example is shown in Fig.~\ref{fig:push_exp_plots}a. The accuracy from 20 simulated tests (10 for each type of trajectory) is 12.3 $\pm$ 4.8\,mm (Table \ref{table:push_eval}, top row). 

\subsubsection{Bi-Reorienting}
In each episode, the object length $l$ is uniformly sampled from $[50,200]$\,mm and a goal angle $\theta^{\rm g}$ is uniformly sampled from $[30^{\circ}, 90^{\circ}]$. Note that the goal angle is evenly divided into 10 subgoals as curriculum learning. The simulated dual-arm tactile robot can learn to reorient various lengths of objects to goal angles with the proposed reward design described in Sec. \ref{subsec:method_task}, with a training curve shown in Fig. \ref{fig:rl_results}b. 
A successful example is shown in Fig.~\ref{fig:reorient_exp_plots}a. 
Note that the translation error is calculated by subtracting the final position of the object from the starting position $p^{\text{o}}_{t=0}$. The average translation and orientation errors from 10 simulated tests are 10.2 $\pm$ 4.8\,mm and $3.4 \pm 1.8^\circ$ respectively (Table \ref{table:orient_eval}).

 \subsubsection{Bi-Gathering}
In each episode, the two object's ${\rm o}^{i}$ ($i=\{1,2\}$) initial positions $p^{{\rm o}_i}_0=(x^i,y^i)$ are uniformly sampled from ranges $x^i\in(-1)^{i}[50,200]$\,mm, $y^i\in(-1)^{i}[0,100]$\,mm, respectively. The other hyperparameters are the termination distance ($\epsilon=90$\,mm; approximate object size), the number of points for subgoals ($n=10$) and updating the subgoal target line every $h=75$ steps. The results (without perturbations) demonstrate that the proposed GUM improves the performance of the simulated dual-arm tactile robot in achieving this task more robustly in shorter episode times (Fig.~\ref{fig:rl_results}c, blue plot), as compared to the one without this mechanism (green plot). This indicates its ability to learn and adapt more effectively: even though the exact object poses are unknown to the robot, it learns to bring together the object locations with only tactile and proprioceptive feedback. 

When perturbations are also considered, these are uniformly sampled from $[1,5]$\,N. In that situation, without GUM, the dual-arm tactile robot cannot learn to achieve this task, with longer episode times compared to the one without perturbations (Fig. \ref{fig:rl_results}c, green and purple plots respectively). With the use of GUM, the task is learned successfully even in the presence of perturbations, reaching a performance similar to that achieved without perturbations (Fig. \ref{fig:rl_results}c, red and blue plots respectively). A successful example with perturbations is shown in Fig.~\ref{fig:gather_exp}a. Upon testing the policy trained under perturbations with the goal-update mechanism, the success rates are 100\% with different perturbation times in each of 5 simulated tests (Table \ref{table:gather_eval}, top row). 
 

\subsection{The Performance of the Bi-Touch in Reality}\label{subsec:real_results}

\subsubsection{Bi-Pushing}

We set up the real bi-pushing task with the same configuration as the simulated one, testing it with three unseen objects (Fig.~\ref{fig:real_objs}a) varying in weight and contact shape. The real dual-arm tactile robot can push a large object through a desired trajectory (example of sinusoidal trajectories in Fig.~\ref{fig:push_exp_plots}b-d). The accuracy from 20 real-world tests (10 for each type of trajectory) for the tripod box, the shuttle tube, and the loudspeaker are 14.2 $\pm$ 6.4\,mm, 16.6 $\pm$ 7.7\,mm, and 17.4 $\pm$ 8.1\,mm respectively (Table~\ref{table:push_eval}), compared to an overall distance travelled of 300-420\,mm. The performances on all objects is similar despite notable differences in their contact shapes (e.g. flat, curved, and sloping surface), showing the generalization ability of the learned policy. Videos for these trajectories are provided in the supplementary materials. 

Comparing the simulated and real trajectories of the end effectors (TacTips), while the box and tube behaved similarly (Fig.~\ref{fig:push_exp_plots}, three left plots), the loudspeaker travelled more distance (right-most plot). We attribute this behaviour to the larger frictional force due to the loudspeaker's heavier weight (879\,g) compared with the box (244\,g) and tube (127\,g). The dual-arm robot needs to move its end-effector closer to the end of the loudspeaker to better counteract the larger frictional force and maintain the object's centre on the goal path. These behaviours show that the dual-arm robot learns to interact with objects of various weights using its tactile feedback, demonstrating the generalization ability and robustness of the learned policy.

\begin{table}
\vspace{-0.5em}
\addtolength{\tabcolsep}{-1pt}
\caption{Mean errors and standard deviations of the object trajectories from the ground truth for the bi-pushing task. The object size is also shown.}
\centering
\begin{tabular}{c|cc} 
\hline
\textbf{Object}         & \textbf{Accuracy} & \textbf{Size}  \\ 
\hline
\textbf{Cuboid (Simulation)}      & 12.3 $\pm$ 4.8~mm           & 400\,mm       \\
\textbf{Box}      & 14.2 $\pm$ 6.4~mm          & 351\,mm        \\
\textbf{Tube}    & 16.6 $\pm$ 7.7~mm            & 386\,mm      \\
\textbf{Loudspeaker}      & 17.4 $\pm$ 8.1~mm     & 377\,mm            \\
\hline
\end{tabular}
\vspace{-1em}
\label{table:push_eval}
\end{table}


\begin{table}
\vspace{-0.5em}
\addtolength{\tabcolsep}{-1pt}
\caption{Mean errors and standard deviations of the reorientation angles and the positions of the objects from the ground truth for the bi-reorienting task.}
\centering
\begin{tabular}{c|ccc} 
\hline
\textbf{Object}         & \textbf{Translation Err.} & \textbf{~Orientation Err.}  & \textbf{Size}  \\ 
\hline
\textbf{Cube (Sim.)}    & 10.2 $\pm $ 4.8\,mm       & 3.4 $\pm $  1.8$^{\circ}$    & 100\,mm           \\
\textbf{Plastic~Cube}                       & 12.5 $\pm $ 5.3\,mm   & 7.5 $\pm $ 3.9$^{\circ}$       & 100\,mm           \\
\textbf{Shoe~Box}                      & 16.2 $\pm $ 6.8\,mm       & 11.5 $\pm $ 5.2$^{\circ}$        & 193\,mm      \\
\textbf{Cracker~Box}                    & 13.9 $\pm $ 5.6\,mm        & 8.4 $\pm $ 4.7$^{\circ}$       & 172\,mm      \\
\textbf{Cubes~Box}                       & 15.5 $\pm $ 5.5\,mm      & 9.2 $\pm $ 5.6$^{\circ}$         & 186\,mm       \\
\textbf{Mustard Bottle}                       & 13.6 $\pm $ 5.4\,mm      & 7.6 $\pm $ 4.1$^{\circ}$         & 122\,mm       \\
\textbf{Goblet}                        & 18.1 $\pm $ 6.1\,mm      & 12.2 $\pm $ 5.8$^{\circ}$        & 113\,mm      \\
\textbf{Soft Brain Toy}                           & 17.2 $\pm $ 6.7\,mm    & 9.7 $\pm $ 5.3$^{\circ}$        & 85\,mm        \\
\textbf{Red~Ball}                      & 19.5 $\pm $ 7.0\,mm      & 13.4 $\pm $ 6.5$^{\circ}$         & 70\,mm     \\
\textbf{Triangular~Prism} & -                  & -           & 57\,mm        \\
\hline
\end{tabular}
\vspace{-1em}
\label{table:orient_eval}
\end{table}

\subsubsection{Bi-Reorienting}
The real bi-reorienting task is considered with the same configuration as the simulated task, testing with a set of new objects that vary in size, shape, weight, and stiffness (Fig.~\ref{fig:real_objs}b). We trained different policies for different directions. Note that the object position is defined as the top centre position where the ArUco marker is attached. The robot is considered to achieve the task when the object angle does not change more than $1^\circ$ in 10 time steps for the final subgoal. The robot achieved this task with most of the selected objects with translation error from 12.5$\pm$5.3\,mm to 19.5$\pm$7.0\,mm, and orientation error from 7.5$\pm$3.9$^{\circ}$ to 13.4$\pm$6.5$^{\circ}$ (Table \ref{table:orient_eval}), except the triangular prism where there was a problem with the sharp edge. Trajectories for a set of real-world objects are shown in Fig.~\ref{fig:reorient_exp_plots} and videos are provided in the supplementary material.

Specifically, the real dual-arm robot can achieve higher accuracy with cube-shaped objects (plastic cube, shoe box, cracker box, mustard bottle, cubes box) compared to round-shaped objects (goblet, plastic ball, soft brain toy). This is because the policy is only trained with cube-shaped objects in simulation so the policy does not have knowledge of the object geometry. We also noticed that the larger the manipulated object is, the larger error will present. We attribute this to the increased difficulty for coordinating both arms to manipulate the larger objects as it requires the end effector to travel more distance during the task. Even so, it was able to successfully manipulate various unseen round objects of various stiffness, which demonstrates the learned policy's generalization ability to rounder objects. On the other hand, we attribute the difficulties on the prism to slippage caused by an unstable contact on the sharper edge, leading to an unrecoverable situation that was never experienced in training.

To further examine the generalization capability, we tried an additional experiment on a set of bricks. Specifically, the bricks are loosely placed together initially, then the robot needs to reorient and gather them together by pushing and rotating the bricks it is contacting. The robot is able to successfully complete this task and, moreover, during the reorientation we can remove the middle brick to demonstrate the robustness to further perturbation. The robot can achieve this (see the video in supplementary material) with a 95\% success rate in 20 tests. Overall, this demonstrates that the policy can lead to a new emergent behaviour not experienced during training.

\begin{figure*}[ht]
  \centering
  \includegraphics[width=0.9\linewidth]{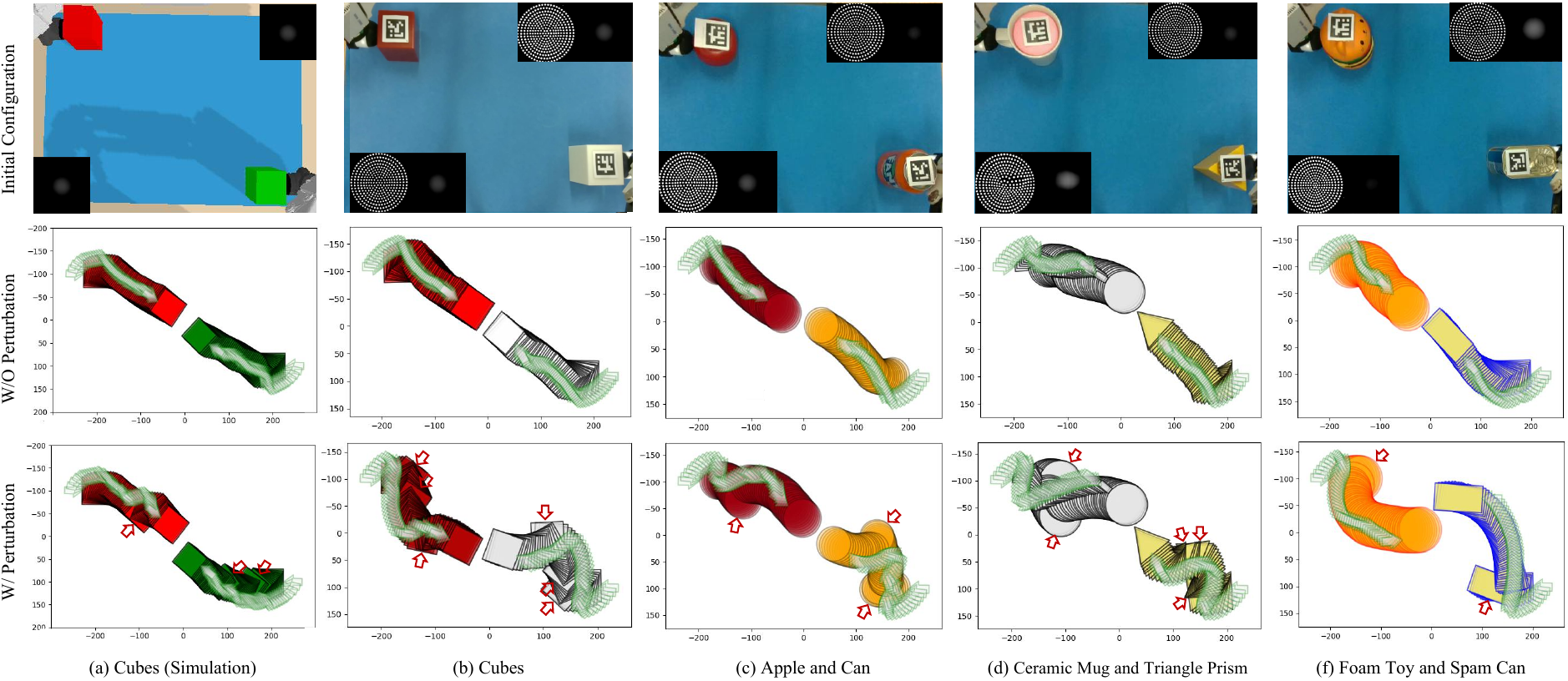}
  \vspace{-1em}
    \caption{Bi-gathering on one simulated pair (a) and four real pairs of objects, corresponding to the (b) cubes, (b) plastic apple and can, (d) ceramic mug and triangular prism and (e) soft foam toy and spam can. The first row shows the initial configurations of the objects and the end effectors (TacTips), along with examples of real and simulated tactile images. The second and third rows show the end-effector (green arrows) and object trajectories. The red arrows in the third row show applied force perturbations. Note that ArUco markers are just for quantitative evaluation, as our policy does not use visual feedback.} 
  \label{fig:gather_exp}
  \vspace{0em}
\end{figure*}

\subsubsection{Bi-Gathering}

The real bi-gathering task is considered with the same configuration as in the simulation, with testing on a set of new objects (Fig.~\ref{fig:real_objs}c) varying in shape, weight and stiffness. Here we applied the policy trained under perturbations in simulation with the target-line goal-update mechanism. We run 10 tests for each object pair with random force perturbations applied at random times. The robot is considered to achieve the task when the centre of the objects becomes closer than 7\,cm (detected using ArUco markers) within 300 time steps. Note that object lengths are about 6\,cm.

The tactile robot successfully completed the bi-gathering task without perturbation in all sets of 10 trials for each object. Regarding the effect of perturbations, the success rates of achieving the task are summarized in Table~\ref{table:gather_eval}. The robot completes the task at 100\% success rate when the perturbation is applied twice or fewer times. The success rate decreases when the number of applied perturbations is increased for all pairs of objects, decreasing most for the irregular items (mug, triangular prism, foam toy and spam can). We show example trajectories of successful trials in Fig.~\ref{fig:gather_exp}. 

Examining the results more closely, the dual-arm tactile robot can complete the task with objects of different weights and shapes to those during training. Comparing the foam toy to the cube in simulation (Figs.~\ref{fig:gather_exp}f, a), the TacTip is deformed much more lightly in reality than in simulation, due to the foam toy's light weight and soft material. Even so, the robot can achieve the task with generated tactile images that are notably different from those in training, both in terms of depression and in the contact shapes (e.g. the mug handle in Fig.~\ref{fig:gather_exp}d). These results demonstrate the strong generalization ability of the policy trained with our goal-update mechanism among different stiffness, shapes and object weights.

We observed that the failure cases are caused by the robot running out of workspace as it tried to push a perturbed object back together through a large turnaround. This happened mostly with the spam can and mug since they present additional challenges for pushing due to their larger weights. 
Even so, the dual-arm tactile robot still demonstrated a capability to push perturbed objects together (Fig. \ref{fig:gather_exp}, third row).

\begin{table}
\addtolength{\tabcolsep}{-4pt}
\caption{Success rates for bi-gathering under different numbers of perturbations.}
\centering
\begin{tabular}{l|l|l|l|l|l|l} 
\hline
\diagbox{\textbf{Objects Pairs}}{\textbf{No. Perturb.}} & 1     & 2     & 3     & 4     & 5    & 6     \\ 
\hline
\textbf{Cube \&~Cube (Simulation)}                                     & 100\% & 100\% & 100\% & 100\% & 100\% & 100\%  \\ 
\hline
\textbf{Cube \&~Cube}                                     & 100\% & 100\% & 100\% & 100\% & 90\% & 90\%  \\ 
\hline
\textbf{Apple \&~Can}                                     & 100\% & 100\% & 100\% & 100\% & 90\% & 70\%  \\ 
\hline
\textbf{Mug \&~Triangular~Prism}                                & 100\% & 100\% & 80\%  & 70\%  & 40\% & 10\%  \\ 
\hline
\textbf{Foam Toy \&~Spam~Can}                               & 100\% & 100\% & 70\%  & 50\%  & 20\% & 10\%  \\
\hline
\end{tabular}
\label{table:gather_eval}
\vspace{-1em}
\end{table}

\section{DISCUSSION AND FUTURE WORK} \label{sec:discussion}

In this paper, we developed a low-cost dual-arm tactile robot system called Bi-Touch for sim-to-real deep reinforcement learning based on Tactile Gym 2.0 \cite{lin2022tactilegym2}. The hardware includes two industry-capable desktop robot arms (Dobot MG400), each equipped with a low-cost high-resolution optical tactile sensor (TacTip) as end-effectors. We also designed a workspace configuration suited for three proposed bimanual tasks tailored towards tactile feedback and integrated into the Tactile Gym simulation methods and environments. 

The performance of our low-cost sim-to-real deep RL dual-arm tactile robot system was evaluated in these three bimanual tasks in the real world. We introduced appropriate reward functions for these tasks in simulation, then investigated how these policies apply to the real world. The experimental results show that the developed dual-arm tactile system is effective for all tasks on real objects unseen in the simulation learning. 


For bi-gathering, we proposed a goal-update mechanism using a linear set of subgoals. These subgoals provide dense auxiliary rewards, enabling the dual-arm robot to learn from a high-dimensional state-action space. While this idea shares similarities with Hierarchical Visual Foresight (HVF) \cite{Nair2020Hierarchical}, we faced challenges in directly applying it to our tactile manipulation tasks due to the limited goal state information provided by tactile images. Instead, we simplified the problem by generating subgoals based on the current object positions in simulation or TCP positions in the real world. This mechanism, which preserves the distribution of initial environmental states, can be seen as a form of implicit curriculum learning \cite{andrychowicz2017hindsight}. It periodically generates goals that are easier to achieve compared to sparse goal settings where only object positions serve as goals. Unlike Hindsight Experience Replay \cite{andrychowicz2017hindsight}, which is applicable only to off-policy RL methods, our method applies on-policy, such as PPO used in this study.


While here we demonstrate the feasibility of our work with 4-DoFs desktop robot arms and TacTips \cite{ward2018tactip}, the proposed framework should also work  with different types of optical tactile sensors, e.g. the GelSight \cite{yuan2017gelsight} and robots with more DoFs, as the Tactile Gym has demonstrated its scalability in both situations \cite{lin2022tactilegym2,church_tactile_2021}. A limitation of the present work is that shear deformation of the tactile sensor has not yet been considered in the simulation and may be needed for more complex tasks relying on the control of frictional forces. A future direction is to approximate the shear effect in simulation with reliable methods to further close the sim-to-real gap. 

A future direction of this work is to apply our Bi-Touch framework and tactile robot platform to fine manipulation of held objects without a supporting table. 
To show the promise of this approach, we additionally developed a bimanual lifting (bi-lifting) task along with a reward function based on principles similar to the other three tasks: a) manipulating the object to achieve a goal state with tactile feedback, b) while maintaining stable contact with the object. Preliminary experiments for the bi-lifting task, both in simulation and the real world, demonstrate that this approach to dexterous manipulation can be effective, which we have included in a supplementary video that accompanies this paper.


\bibliographystyle{unsrt}
\bibliography{manual}

\end{document}

%% file: math/rap2texdefs.tex
\usepackage{amsmath,amsfonts,amssymb}

\newcommand{\beq}{\begin{equation}}
\newcommand{\eeq}{\end{equation}}
\newcommand{\bear}{\begin{eqnarray}}
\newcommand{\bears}{\begin{eqnarray*}}
\newcommand{\eear}{\end{eqnarray}}
\newcommand{\eears}{\end{eqnarray*}}
\newcommand{\bdm}{\begin{displaymath}}
\newcommand{\edm}{\end{displaymath}}
\newcommand{\lba}{\left[\begin{array}}
\newcommand{\ear}{\end{array}\right]}